\documentclass[12pt]{article}

\usepackage[margin=1in]{geometry}

\usepackage{setspace}

\usepackage{graphicx}

\usepackage{enumitem}

\usepackage{microtype}
\usepackage{subfigure}
\usepackage{booktabs}
\usepackage{comment}
\usepackage{wrapfig}
\usepackage{arydshln}
\usepackage{enumitem}
\usepackage{multirow}
\usepackage{url}
\usepackage{natbib}
\usepackage{authblk}

\RequirePackage[colorlinks,citecolor=blue,linkcolor=blue,urlcolor=blue,pagebackref]{hyperref}

\usepackage{amsmath}
\usepackage{amssymb}
\usepackage{mathtools}
\usepackage{amsfonts}
\usepackage{bm}
\usepackage{bbm}

\usepackage{algorithm}
\usepackage{algorithmic}

\def\eqref#1{equation~\ref{#1}}

\def\1{\bm{1}}

\DeclareMathAlphabet{\mathsfit}{\encodingdefault}{\sfdefault}{m}{sl}
\SetMathAlphabet{\mathsfit}{bold}{\encodingdefault}{\sfdefault}{bx}{n}

\def\calF{{\mathcal{F}}}
\def\calG{{\mathcal{G}}}

\def\calX{{\mathcal{X}}}

\def\bbR{{\mathbb{R}}}

\DeclareMathOperator*{\argmax}{arg\,max}
\DeclareMathOperator*{\argmin}{arg\,min}

\newcommand{\p}[1]{\left(#1\right)}
\newcommand{\sqb}[1]{\left[#1\right]}
\newcommand{\cb}[1]{\left\{#1\right\}}

\newcommand{\Bigp}[1]{\Big(#1\Big)}
\newcommand{\Bigsqb}[1]{\Big[#1\Big]}
\newcommand{\Bigcb}[1]{\Big\{#1\Big\}}

\newcommand{\bigp}[1]{\big(#1\big)}

\usepackage{amsthm}

\theoremstyle{plain}

\newtheorem{theorem}{Theorem}[section]

\usepackage[textsize=tiny]{todonotes}
\usepackage{multirow}
\usepackage{wrapfig}
\usepackage{subfigure}

\renewcommand{\eqref}[1]{(\ref{#1})}

\usepackage{xcolor}
\newcount\Comments  
\newcommand{\kibitz}[2]{\ifnum\Comments=1\textcolor{#1}{#2}\fi}

\usepackage[capitalize,noabbrev]{cleveref}

\allowdisplaybreaks

\title{Bridging the Gap between Empirical Welfare Maximization and Conditional Average Treatment Effect Estimation in Policy Learning}

\author{Masahiro Kato\thanks{Email: \texttt{mkato-csecon@g.ecc.u-tokyo.ac.jp}}$\,$}

\affil{The University of Tokyo}

\date{\today}

\begin{document}

\maketitle 

\begin{abstract}
The goal of policy learning is to train a policy function that recommends a treatment given covariates to maximize population welfare. There are two major approaches in policy learning: the empirical welfare maximization (EWM) approach and the plug-in approach. The EWM approach is analogous to a classification problem, where one first builds an estimator of the population welfare, which is a functional of policy functions, and then trains a policy by maximizing the estimated welfare. In contrast, the plug-in approach is based on regression, where one first estimates the conditional average treatment effect (CATE) and then recommends the treatment with the highest estimated outcome. This study bridges the gap between the two approaches by showing that both are based on essentially the same optimization problem. In particular, we prove an exact equivalence between EWM and least squares over a reparameterization of the policy class. As a consequence, the two approaches are interchangeable in several respects and share the same theoretical guarantees under common conditions. Leveraging this equivalence, we propose a regularization method for policy learning. The reduction to least squares yields a smooth surrogate that is typically easier to optimize in practice. At the same time, for many natural policy classes the inherent combinatorial hardness of exact EWM generally remains, so the reduction should be viewed as an optimization aid rather than a universal bypass of NP-hardness.
\end{abstract}

\section{Introduction}
Decision making about treatment choice is a central objective in causal inference \citep{Manski2002treatmentchoice}. In this study, to recommend a treatment for an individual with covariates $X$, we aim to train a \emph{policy function} that maps covariates to treatment recommendations using observational data. This \emph{policy learning} problem has been widely studied in economics, statistics, and machine learning \citep{Swaminathan2015counterfactualrisk,Swaminathan2015batchlearning,Kitagawa2018whoshould,Athey2021policylearning,Zhou2023offlinemultiaction}.

There are two main approaches in policy learning. The first is \emph{empirical welfare maximization} (EWM) or equivalently, counterfactual risk minimization, where we estimate the welfare for each candidate policy and select the one that maximizes the estimated welfare \citep{Swaminathan2015counterfactualrisk,Swaminathan2015batchlearning,Kitagawa2018whoshould}. The second is the \emph{plug-in} approach, where we estimate the conditional average treatment effect (CATE) and treat whenever the estimated effect is nonnegative. Existing studies have argued that the EWM approach is more preferable, since it directly targets the policy objective rather than relying on an intermediate regression step. For instance, from a theoretical viewpoint, \citet{Kitagawa2018whoshould} derive regret bounds for EWM with VC-type policy classes, yielding $1/\sqrt{n}$-type rates for the minimax regret. Note that such (minimax) optimal rates depend on the type of worst-case scenarios considered, as discussed in \citet{Audibert2007fastlearning}. When the underlying expected outcomes belong to a H\"{o}lder class, the minimax optimal rate is typically characterized by the nonparametric rate $n^{\beta/(2\beta + d)}$. However, if we apply VC-type bounds, we cannot achieve this rate, since the VC dimension of the H\"{o}lder class is infinite.

We reconsider the distinction between the EWM and plug-in approaches and aim to bridge the gap between them. Our key message is that the two are \emph{two faces of the same underlying optimization}: EWM can be reformulated as least squares on the CATE with certain restrictions on the regression models. This observation has two important practical implications. First, it conceptually unifies the literature by showing that EWM and plug-in are equivalent under restricted regression models. Second, it allows us to avoid the NP-hard combinatorial optimization that arises in the EWM approach.

\paragraph{Contributions.}
Here, we summarize our contributions:
\begin{enumerate}
    \item \textbf{Equivalence.} We establish that EWM over a policy class $\Pi$ is equivalent to least squares over the class $\calG_\Pi\coloneqq\{g=2\pi-1:\pi\in\Pi\}$ for the target $Y_1-Y_0$. At the empirical level, EWM with an inverse-probability-weighted welfare estimator equals least squares of an IPW pseudo-outcome on $g\in\calG_\Pi$, if the propensity score is known.
    \item \textbf{Computation.} The equivalence provides a least–squares training objective whose continuous form enables standard optimization tools and simplifies implementation. For common discrete policy classes (e.g., trees, rule lists, set indicators), this surrogate often eases training in practice, but it does not in itself remove the intrinsic combinatorial hardness of exact EWM.
\end{enumerate}

\section{Setup}
We mostly follow \citet{Kitagawa2018whoshould} while allowing for both deterministic ($0$–$1$) and randomized ($[0,1]$) policies.

\paragraph{Observations.}
Let the sample size be $n$. We observe i.i.d.\ draws $Z_i=(Y_i,D_i,X_i)$, $i=1,\dots,n$, where $X_i\in\calX\subset\bbR^{d_\calX}$ are pre-treatment covariates, $D_i\in\{0,1\}$ is a binary treatment indicator, and $Y_i\in\bbR$ is the observed outcome.

\paragraph{Potential outcomes.}
Under the Neyman–Rubin model, each unit has potential outcomes $(Y_{0,i},Y_{1,i})$ and $Y_i=D_i Y_{1,i}+(1-D_i)Y_{0,i}$. We assume \emph{unconfoundedness}, $(Y_{0,i},Y_{1,i})\perp\perp D_i \mid X_i$, and define $P$ as the population distribution of $(Y_{0},Y_{1},D,X)$.

\paragraph{Policy.}
Let $\Pi$ be a class of policies $\pi:\calX\to[0,1]$. In \citet{Kitagawa2018whoshould}, a common deterministic class is $\pi(X)=\mathbbm{1}\{X\in G\}$ for $G\subset\calX$. We analyze both $0$–$1$ and $[0,1]$ policies.

\paragraph{Welfare.}
The (utilitarian) social welfare of $\pi$ is
\[
W(\pi)\coloneqq E_P\bigl[Y_1\pi(X)+Y_0(1-\pi(X))\bigr].
\]
This reduces to $W(G)=E_P[Y_1\mathbbm{1}\{X\in G\}+Y_0\mathbbm{1}\{X\notin G\}]$ when $\pi=\mathbbm{1}\{X\in G\}$.

\paragraph{Goal and regret.}
Given data, we construct $\widehat\pi$ and assess it via regret
\[
\mathrm{Regret}\bigl(\pi^*_\Pi,\widehat\pi\bigr)\coloneqq W(\pi^*_\Pi)-W(\widehat\pi),
\qquad
\pi^*_\Pi \in \argmax_{\pi\in\Pi} W(\pi).
\]
Let $\tau(x)\coloneqq E[Y_1-Y_0\mid X=x]$ denote the CATE. The first-best policy over all measurable policies is
\[
\pi^*_{\mathrm{FB}}(x)=\mathbbm{1}\{\tau(x)\ge 0\}.
\]
If $\Pi$ is unrestricted, then $\pi^*_\Pi=\pi^*_{\mathrm{FB}}$; otherwise, $\pi^*_\Pi$ is the second-best policy within $\Pi$.

\subsection*{Notation and assumptions}
Let $e(x)=P(D=1\mid X=x)$ be the propensity score and $m_d(x)=E[Y_d\mid X=x]$ the conditional mean outcomes. We assume that there exists a constant $0< \epsilon< 1/2$ independent of $n$ such that $\epsilon < e(X) < 1 - \epsilon$ almost surely and that the variables $X$ and $Y$ are bounded. 

\section{Recap of EWM and plug-in approaches}

\subsection{EWM Policy}
Given an estimator $\widehat W(\pi)$ of $W(\pi)$, we train a policy as
\[
\widehat\pi_{\mathrm{EWM}} \in \argmax_{\pi\in\Pi} \widehat W(\pi).
\]
A basic estimator of the welfare $W(\pi)$ is the IPW estimator, defined as
\[
\widehat W_n^{\mathrm{IPW}}(\pi)
\coloneqq \frac{1}{n}\sum_{i=1}^n\p{\frac{Y_i D_i}{e(X_i)}\pi(X_i)+\frac{Y_i(1-D_i)}{1-e(X_i)}\bigp{1-\pi(X_i)}}.
\]

When $e(\cdot)$ is unknown, we use an estimate of $\widehat{e}(\cdot)$ instead of its true value. To remove estimation bias at a fast rate, we can employ the augmented IPW (AIPW), also called the doubly robust (DR), welfare estimator, defined as
\begin{align*}
    &\widehat W_n^{\mathrm{DR}}(\pi) \coloneqq\\
    &\frac{1}{n}\sum_{i=1}^n\p{
\pi(X_i)\p{\frac{D_i}{\widehat{e}(X_i)}\bigp{Y_i-\widehat{m}_1(X_i)} + \widehat{m}_1(X_i)} + (1 - \pi(X_i))\p{\frac{1 - D_i}{1-\widehat{e}(X_i)}\bigp{Y_i-\widehat{m}_0(X_i)} + \widehat{m}_0(X_i)}},
\end{align*}
where $\widehat{e}(x)$ is an estimator of the propensity score $e(x)$, and $\widehat{m}_d(x)$ ($d\in \{1, 0\}$) is an estimator of the conditional expected outcome $m_d(x)$ \citep{Athey2021policylearning}. If the estimators of the nuisance parameters satisfy the Donsker condition or are constructed via sample splitting \citep{Klaassen1987consistentestimation}, also called cross-fitting \citep{Chernozhukov2018doubledebiased}, and meet mild convergence rate conditions, then the bias from estimation error vanishes at a faster rate than $1/\sqrt{n}$. We can estimate the bias-correction weight by using Riesz regression or density-ratio estimation, as proposed in \citet{Chernozhukov2024automaticdebiased,Chernozhukov2022debiasedmachine,Kato2025directbias,Kato2025directdebiased}.

\subsection{Plug-in Policy}
The plug-in approach first estimates the CATE $\tau(X)$. We can estimate $\tau(X)$ by estimating the conditional expected outcomes $m_d(X) = E_P[Y_d\mid X]$. Let $\widehat{\tau}(X)$ and $\widehat{m}_d(X)$ be estimators of $\tau(X)$ and $m_d(X)$, respectively. For example, we can estimate $m_d(X)$ by regressing $Y_i$ on $X_i$ using only data with $D_i = d$; that is, we estimate $m_d$ as
\[\widehat{m}_d \coloneqq \argmin_{f\in\calF} \sum^n_{i=1}\mathbbm{1}[D_i = d]\bigp{Y_i - f(X_i)}^2,\]
where $\calF$ is a class of regression functions $f\colon \calX \to \bbR$, and construct $\widehat{\tau}$ as $\widehat{\tau} = \widehat{m}_1 - \widehat{m}_0$. 
As another example, we can estimate $\tau(X)$ directly by regressing a variable that is (asymptotically) unbiased for $Y_{1, i} - Y_{0, i}$ on covariates $X_i$. For instance, using an IPW estimator for the target variable, we estimate $\tau$ as
\[\widehat{\tau} \coloneqq \argmin_{f\in\calF} \sum^n_{i=1}\p{\frac{Y_i D_i}{e(X_i)} - \frac{Y_i (1 - D_i)}{1 - e(X_i)} - f(X_i)}^2,\]

Using an estimator $\widehat{m}_d(X)$, we construct a policy as
\begin{align*}
    \widehat{\pi}_{\text{plug}\mathchar`-{in}}(X) = \begin{cases}
        1 & \text{if}\ \ \widehat{\tau}(X) \geq 0\\
        0 & \text{if}\ \ \widehat{\tau}(X) < 0
    \end{cases}. 
\end{align*}

\section{Equivalence between EWM and least squares}
We now formalize the core equivalence. Throughout this section, let $\calG_\Pi\coloneqq\{g=2\pi-1:\pi\in\Pi\}$.

\subsection{$0$–$1$ policies}
We first consider the case where $\pi$ is a $0$-$1$-valued policy; that is, $\pi \colon \calX \to \{0, 1\}$, where $\pi(X) = 1$ (resp.\ $\pi(X) = 0)$ means that the policy recommends treatment $1$ (resp.\ treatment $0$) for given covariates $X$.

\paragraph{Oracle problem.} We show the equivalence between the EWM and plug-in approaches when we can observe $Y_1$ and $Y_0$ directly, ignoring the counterfactual nature. 
Recall that the welfare maximization problem can be written as
\begin{align*}
    \max_{\pi\in\Pi} E_P\sqb{\pi(X)Y_1 + (1 - \pi(X))Y_0}. 
\end{align*}

In this case, we consider the following regression problem:
\begin{align*}
    &\min_{g \in \calG_\Pi} E_P\sqb{\Bigp{\bigp{Y_1 - Y_0} - g(X)}^2},
\end{align*}
where $\calG_\Pi \coloneqq \{g = (2\pi - 1) \colon \pi \in \Pi\}$. By definition, $g$ is a function such that $g\colon \calX \to \{-1, 1\}$.

We now show the equivalence between EWM and least squares. 
Recall that we defined the optimal policy $\pi^*$ as
\begin{align*}
    \pi^* = \argmax_{\pi\in\Pi} W(\pi) = \argmax_{\pi\in\Pi} E_P\sqb{\pi(X)Y_1 + (1 - \pi(X))Y_0}.
\end{align*}

Let $g^*$ be the optimal predictor defined as
\begin{align*}
    g^* \coloneqq \argmin_{g \in \calG_\Pi}E_P\sqb{\Bigp{\bigp{Y_1 - Y_0} - g(X)}^2}.
\end{align*}

Then, the following theorem holds.
\begin{theorem}
\label{thm:popequivalence}
It holds that
\[g^* = 2\pi^* - 1.\]
\end{theorem}
The proof is straightforward and shown below.

\begin{proof}
For $\calG_\Pi$, we have
\begin{align*}
    g^* = &\argmin_{g \in \calG_\Pi}E_P\sqb{\Bigp{\bigp{Y_1 - Y_0} - g(X)}^2}\\
    &= \argmin_{g \in \calG_\Pi}E_P\sqb{\bigp{Y_1 - Y_0}^2 - 2g(X)\bigp{Y_1 - Y_0} + g(X)^2}\\
    &= \argmin_{g \in \calG_\Pi}E_P\sqb{ - 2g(X)\bigp{Y_1 - Y_0} + g(X)^2}\\
    &= \argmin_{g \in \calG_\Pi}E_P\sqb{ - 2g(X)\bigp{Y_1 - Y_0} + 1}\\
    &= \argmin_{g \in \calG_\Pi}E_P\sqb{ - 2g(X)\bigp{Y_1 - Y_0} }.
\end{align*}
Here, we used $g(X)^2 = 1$. We omitted $\bigp{Y_1 - Y_0}^2$ from the second to third line and $1$ from the fourth to fifth, since they are irrelevant to the optimization. 

Continuing, we have
\begin{align*}
    g^* = &\argmin_{g \in \calG_\Pi}E_P\sqb{ - 2g(X)\bigp{Y_1 - Y_0} }\\
    &= \argmin_{g \in \calG_\Pi}E_P\sqb{ - 2\bigp{g(X) + 1}\bigp{Y_1 - Y_0} + 2 \bigp{Y_1 - Y_0}}.
\end{align*}
We added and subtracted terms that are irrelevant to the optimization. Here, recall that 
\begin{align*}
    \pi^* &= \argmin_{\pi \in \cb{\pi(\cdot) = \bigp{g(\cdot) + 1}/2 \colon g \in \calG_\Pi}}E_P\sqb{ - \pi(X)Y_1 - (1 - \pi(X))Y_0}\\
    &= \argmin_{\pi \in \cb{\pi(\cdot) = \bigp{g(\cdot) + 1}/2 \colon g \in \calG_\Pi}}E_P\sqb{ - \pi(X)(Y_1 - Y_0) - Y_0}.
\end{align*}
Therefore, $g^* = 2\pi^* - 1$ holds, and the proof is complete.
\end{proof}

This theorem implies that the EWM approach is equivalent to least squares, where we regress $Y_1 - Y_0$ using a function $g \colon \calX \to \{-1, 1\}$.

\paragraph{Empirical version.}
We can similarly show the empirical version of this equivalence. Recall that given true $e(x)$, the EWM approach trains a policy $\widehat{\pi}_n$ as
\begin{align*}
    \widehat{\pi} \coloneqq \argmax_{\pi\in\Pi} W_n(\pi) = \argmax_{\pi\in\Pi} \frac{1}{n}\sum^n_{i=1}\p{\frac{Y_i D_i}{e(X_i)}\pi(X_i) + \frac{Y_i (1 - D_i)}{1 - e(X_i)}(1 - \pi(X_i)) }.
\end{align*}
Let $\widehat{g}_n$ be the predictor for the CATE defined as
\begin{align*}
    \widehat{g} \coloneqq \argmin_{g \in \calG_\Pi} \frac{1}{n}\sum^n_{i=1}\p{\frac{Y_i D_i}{e(X_i)} - \frac{Y_i (1 - D_i)}{1 - e(X_i)} - g(X_i)}^2.
\end{align*}
Then, the following holds:
\begin{theorem}
It holds that
\[\widehat{g} = 2\widehat{\pi} - 1.\]
\end{theorem}
The proof follows the same logic as Theorem~\ref{thm:popequivalence} and is omitted for brevity.

\subsection{$[0, 1]$-policy}
We next consider the case where $\pi$ is a $[0, 1]$-valued function; that is, $\pi\colon \calX \to [0, 1]$. Let $\Pi$ be a policy class that contains such policies, $\pi\colon \calX \to [0, 1]$.

\paragraph{Oracle problem}
We consider the equivalence between the following two problems: regularized EWM and modified least squares. We define the regularized EWM as follows:
\begin{align*}
\max_{\pi\in\Pi} \cb{W(\pi) - \lambda E_P\Bigsqb{\bigp{2\pi(X) - 1}^2}},
\end{align*}
where $\lambda > 0$ is a regularization coefficient. Note that if $\pi(X) = 1$, then $2\pi(X) - 1 = 1$; if $\pi(X) = 0$, then $2\pi(X) - 1 = -1$. This regularization term helps prevent the policy from overfitting to the observations by discouraging extreme values of $1$ or $0$.

We also define a corresponding least squares problem. Given a regularization coefficient $\lambda > 0$, we define the mean squared error between $\frac{1}{\sqrt{\lambda}} \bigp{Y_1 - Y_0}$ and $\sqrt{\lambda} g(X)$ as
\begin{align*}
&\min_{g \in \calG_\Pi}E_P\sqb{\p{\frac{1}{\sqrt{\lambda}} \bigp{Y_1 - Y_0} - \sqrt{\lambda} g(X)}^2}.
\end{align*}

Let $\widetilde{\pi}(\lambda)$ be an optimal policy defined as
\begin{align*}
\widetilde{\pi}(\lambda) \coloneqq \argmax_{\pi\in\Pi} \cb{W(\pi) - \lambda E_P\Bigsqb{\bigp{2\pi(X) - 1}^2}}.
\end{align*}

Let $\widetilde{g}(\lambda)$ be an optimal predictor defined as
\begin{align*}
\widetilde{g}(\lambda) \coloneqq \argmin_{g \in \calG_\Pi}E_P\sqb{\p{\frac{1}{\sqrt{\lambda}} \bigp{Y_1 - Y_0} - \sqrt{\lambda} g(X)}^2}.
\end{align*}

We now show the equivalence between these EWM and least squares formulations.

\begin{theorem}
For any $\lambda > 0$, it holds that
\[\widetilde{g}(\lambda) = 2\widetilde{\pi}(\lambda / 4) - 1.\]
In addition, as $\lambda \to 0$, we have
\[\widetilde{\pi}(\lambda) \to \pi^*,\]
which also implies $\widetilde{g}(\lambda) \to 2 \pi^* - 1$ as $\lambda \to 0$.
\end{theorem}

The proof is straightforward and shown below.

\begin{proof}
For $\calG_\Pi$, we have
\begin{align*}
\widetilde{g}(\lambda)&= \argmin_{g \in \calG_\Pi}E_P\sqb{\p{\frac{1}{\sqrt{\lambda}} \bigp{Y_1 - Y_0} - \sqrt{\lambda} g(X)}^2}\\
&= \argmin_{g \in \calG_\Pi}E_P\sqb{\frac{1}{\lambda}\bigp{Y_1 - Y_0}^2 - 2g(X)\bigp{Y_1 - Y_0} + \lambda g(X)^2}\\
&= \argmin_{g \in \calG_\Pi}E_P\sqb{ - 2g(X)\bigp{Y_1 - Y_0} + \lambda g(X)^2}\\
&= \argmin_{g \in \calG_\Pi}E_P\sqb{ - 2\bigp{g(X) + 1}\bigp{Y_1 - Y_0} + 2 \bigp{Y_1 - Y_0} + \lambda g(X)^2}\\
&= \argmin_{g \in \calG_\Pi}\Bigcb{E_P\sqb{ - \bigp{g(X) + 1}\bigp{Y_1 - Y_0} / 2} + \lambda E_P\sqb{ g(X)^2} / 4}.
\end{align*}
We also have
\[\widetilde{\pi}(\lambda) = \argmin_{\pi \in \cb{\pi(\cdot) = \bigp{g(\cdot) + 1}/2 \colon g \in \calG_\Pi}}\Bigcb{E_P\sqb{ - \pi(X)Y_1 - (1 - \pi(X))Y_0} + \lambda E_P\sqb{ \bigp{2\pi(X) - 1}^2}}.\]
Since $g = 2 \pi - 1$, the proof completes. 
\end{proof}

\paragraph{Empirical version}
Similarly, we can show the empirical version of the equivalence result. In empirical analysis, we can only observe either $Y_1$ or $Y_0$ based on the treatment $D$. Therefore, we need to estimate $(Y_1-Y_0)$ as a pseudo-outcome.

Given known $e(\cdot)$, we define a policy empirically trained with regularized EWM as
\begin{align*}
\widehat{\pi}(\lambda) \coloneqq \argmax_{\pi\in\Pi} \cb{\widehat{W}_n^{\mathrm{IPW}}(\pi) - \lambda \frac{1}{n}\sum^n_{i=1}\bigp{2\pi(X_i) - 1}^2}.
\end{align*}

We also define a trained predictor $\widehat{g}_n$ for the CATE as
\begin{align*}
\widehat{g}(\lambda) \coloneqq \argmin_{g \in \calG_\Pi} \frac{1}{n}\sum^n_{i=1}\p{\frac{1}{\sqrt{\lambda}}\p{\frac{Y_i D_i}{e(X_i)} - \frac{Y_i (1 - D_i)}{1 - e(X_i)}} - \sqrt{\lambda}g(X_i)}^2.
\end{align*}

Then, the following theorem holds. Since the proof is almost the same as that of the above theorem, we omit it.

\begin{theorem}
For any $\lambda > 0$, it holds that
\[\widehat{g}(\lambda) = 2\widehat{\pi}(\lambda / 4) - 1.\]
In addition, as $\lambda \to 0$, we have
\[\widehat{\pi}(\lambda) \to \widehat{\pi},\]
which also implies $\widehat{g}(\lambda) \to 2 \widehat{\pi} - 1$ as $\lambda \to 0$.
\end{theorem}

\section{Computational implications}
Maximizing $\widehat{W}(\pi)$ over $\pi\in\Pi$ is, in general, a combinatorial problem (e.g., empirical set selection) and is NP–hard for many natural classes $\Pi$. The equivalence established above replaces this discrete search with empirical risk minimization under a squared loss on a pseudo–outcome, followed by the simple back–mapping $g\mapsto \pi=(g+1)/2$. This yields a training pipeline that is substantially easier to optimize than direct $0$–$1$ search.

The learning problem reduces to least squares regression of a pseudo–outcome, such as an IPW estimator, on a predictor $g\in\mathcal G_\Pi=\{2\pi-1:\pi\in\Pi\}$. This reformulation enables the use of standard continuous optimization tools (closed–form solvers, SGD, or second–order methods).

If the model class for $g$ is convex, then the least–squares is a convex optimization problem. In these cases, global optima are attained by off–the–shelf solvers. When $\Pi$ is discrete/nonconvex (e.g., trees, rule lists, or set indicators), the induced class $\mathcal G_\Pi$ is also nonconvex, so the objective need not be convex. Even then, the squared–loss formulation remains markedly easier to handle than maximizing a combinatorial welfare objective, because it replaces $0$–$1$ structure with a smooth surrogate amenable to gradient–based methods. 

Regardless of the specific function class, the least–squares view avoids the NP–hard policy search that arises in direct EWM and provides a stable, scalable ERM workflow with standard regularization and tuning.

\section{Conclusion}
From the decision-making, EWM and plug-in policies are often presented as distinct. From a decision-making perspective, the former optimizes welfare directly, whereas the latter bypasses the estimation of treatment effects as an intermediate step. We showed that this distinction collapses under a simple reparameterization: maximizing estimated welfare over a policy class is equivalent to least squares on a CATE target with predictors constrained to $g=2\pi-1$. This equivalence clarifies the relationship between the two traditions, allows regression guarantees to inform policy learning, and replaces combinatorial search with an empirical risk minimization routine that is typically easier to optimize in practice, while the underlying combinatorial hardness of exact policy search generally remains.

The same viewpoint extends beyond binary treatments. For multiple treatments, one can work with one-vs-all pseudo-outcomes or a simplex-valued predictor and a squared-loss on a vector CATE. Budget or fairness constraints can be incorporated through penalties in the least-squares objective. We hope this unification simplifies both analysis and implementation of policy learning in applied work, and we anticipate that remaining gaps in theoretical results can be narrowed under common regularity conditions.

\bibliography{arXiv2.bbl}
\bibliographystyle{tmlr}

\end{document}